\documentclass[runningheads]{llncs}
\usepackage[T1]{fontenc}
\usepackage{graphicx}
\usepackage{booktabs}
\usepackage[misc]{ifsym}
\newcommand{\corr}{(\Letter)}
\usepackage{mwe}
\usepackage{url}            
\usepackage{xcolor}         
\usepackage{amsmath,amssymb,amsfonts}
\usepackage{tabularx}
\usepackage{bbm}
\usepackage{subfig}
\usepackage[acronym, shortcuts]{glossaries}

\newcommand{\best}[1]{\pdfliteral direct {2 Tr 0.3 w}#1\pdfliteral direct {0 Tr 0 w}} 
\newcommand{\secondbest}[1]{{\underline{{#1}}}}

\newcolumntype{L}{>{\raggedright\arraybackslash}X}
\newcolumntype{C}{>{\centering\arraybackslash}X}
\newcolumntype{R}{>{\raggedleft\arraybackslash}X}

\newcommand{\mat}[1]{\mathbf{\uppercase{#1}}} 
\newcommand{\myvec}[1]{{\mathbf{\lowercase{#1}}} }

\newcommand{\letterfunc}[1]{\mathcal{#1}} 
\newcommand{\myset}[1]{\mathbb{#1}}

\newacronym{eo}{EO}{Earth Observation}
\newacronym{f1}{F1}{weighted F1}
\newacronym{moddrop}{ModDrop}{Modality Dropout}
\newacronym{fullco}{FullCo}{Full Co-learning}
\newacronym{comiss}{Co-Miss}{Co-learning for Missing}
\newacronym{geofm}{GeoFM}{Geospatial Foundation Models}

\begin{document}

\title{Co-Learning for Missing Arbitrary Modalities in Multi-modal Classification}

\author{Francisco Mena\inst{1}\orcidID{0000-0002-5004-6571}\corr \and
Dino Ienco\inst{2,3}\orcidID{0000-0002-8736-3132} \and
Roberto Interdonato\inst{4,3}\orcidID{0000-0002-0536-6277} \and
Cassio F. Dantas\inst{2,3}\orcidID{0000-0002-1934-0625} \and 
Simon Besnard \inst{1}\orcidID{0000-0002-1137-103X}}

\authorrunning{Mena et al.}

\institute{GFZ Helmholtz Center for Geosciences, Potsdam, Germany \email{mena@gfz.de}
\and
INRAE, UMR TETIS, University of Montpellier, Montpellier, France 
\and
INRIA, EVERGREEN, University of Montpellier, Montpellier, France 
\and
CIRAD, UMR TETIS, University of Montpellier, Montpellier, France
}

\maketitle              

\begin{abstract} 
Multi-modal classification leverages complementary information across diverse data sources to enhance predictive performance.
However, real-world scenarios subject to operational constraints, such as sensor failures or privacy restrictions, lead to inconsistent modality availability between training and inference times. 
To handle missing modalities, prior studies have mainly covered bimodal data setups and focused on designing robust fusion processes.
Instead, we adopt a multi-modal co-learning framework that prioritizes inter-modal collaboration rather than multi-modal fusion. 
Specifically, we consider that any subset of modalities may be absent, without assuming predefined missing-modality patterns, an inference scenario we refer to as missing arbitrary modalities.
To address this challenge, we introduce two alternative approaches that leverage information at both feature- and decision-level.
Experiments on two multi-modal classification benchmarks demonstrate significant robustness gains in various missing modality conditions. 
The first method shows more robust behavior under minimal missing conditions, where a single modality is absent, whereas the second performs better under extreme missing conditions, where all-but-one modalities are missing.
\textbf{Our code is available at \url{https://github.com/fmenat/Co4Miss}.}
\keywords{Multi-modal classification \and Co-learning \and Missing modalities \and Robustness.}
\end{abstract}

\section{Introduction} \label{sec:intro}

Multi-modal classification aims to fuse complementary data sources to enhance predictive performance and support reliable decision-making \cite{yan2021deep}. 
Although related research primarily focuses on standard multi-modal setups such as image–text–audio \cite{yan2021deep}, real-world applications often involve heterogeneous modalities beyond this setup. 
Nowadays, modern sensing technologies collect a vast amount of data modalities with varying characteristics, e.g. acquisition processes, resolutions, and physical meaning. 
In the \gls{eo} domain, satellite imagery, radar, LiDAR, and in-situ measurements provide structurally distinct yet complementary views of the Earth's surface. 
In human-centered studies, wearable and biometric sensors produce diverse physiological and motion signals of a subject. 
Whereas such diversity enriches representation, it also poses significant challenges if training modalities are absent at inference time.

In real-world applications, data modalities cannot be assumed to be consistently accessible across the same phenomenon and time span.
This is because data acquisition can be constrained by environmental, operational, and cost factors. 
As a result, modalities used at training time may be partially observed or entirely missing at inference \cite{wu2024deep}. 
This challenge is evident in \gls{eo}, where optical imagery may be affected by cloud coverage, satellite missions can be discontinued, or may operate only over restricted geographic areas, leading to abrupt gaps in multimodal data sources~\cite{shen2015missing}. 
Similar issues arise in human-centered systems relying on wearable sensors, where motion signals may be corrupted, interrupted by battery depletion, or unavailable due to user non-compliance. 
These situations underscore the need for multi-modal models capable of remaining robust when one or more modalities are missing at deployment.

The challenge of missing modalities has been addressed with simple data processing techniques (e.g. imputation) to more advanced DL approaches \cite{wu2024deep}.
Most DL methods enforce the fusion process to be robust, e.g. by randomly dropping modalities during training.
This \gls{moddrop} technique \cite{neverova2015moddrop} can be implemented in various ways in the literature.
For instance, data modalities are replaced with zero at input- \cite{neverova2015moddrop}, or feature-levels \cite{lin2023missmodal}, filled up with a learnable parameter \cite{wu2024deep}  
masked out from the attention \cite{ma2022multimodal}, or completely ignored during fusion \cite{mena2025maug}. 
To recover the full set of training modalities before the fusion time, models like SMIL \cite{ma2021smil}, and ActionMAE \cite{woo2023towards} perform cross-modal reconstruction.
Beyond designing robust fusion processes, Ma et al. \cite{ma2022multimodal} introduce a sharing weights mechanism across modality-dedicated components.
Moreover, McKinzie et al. \cite{mckinzie2023robustness} harness the knowledge distillation framework with a full-modal teacher who guides a student with partial modalities available.

Recently, the multi-modal co-learning framework has emerged for addressing the challenge of missing modalities.
In the co-learning paradigm, multiple models are trained to cooperate (share knowledge), aiming to enhance their individual performance \cite{rahate2022multimodal}. 
Beyond its common usages for domain adaptation, 
noisy labels, 
and knowledge distillation, 
few works have harnessed it in multi-modal data setups. 
This involves the collaboration between modality-dedicated models or components, either in a model-, feature-, or decision-based approach.
For instance, MLA~\cite{zhang2024multimodal} shares the last layers of modality-specific models (model-based) and adjusts gradient directions to avoid overwriting modality-specific knowledge.
MDiCo \cite{mena2025mdico} enforces the learning of shared and specific features in modality-dedicated models (feature-based) for unimodal boosting in a bimodal setup. 
MV-HFMD \cite{black2024multi} uses mutual distillation, previously introduced in \cite{zhang2018deep}, to guide individual per-modality predictions toward a full-modality consensus.

\begin{figure}[!t]
    \centering
    \includegraphics[width=\linewidth]{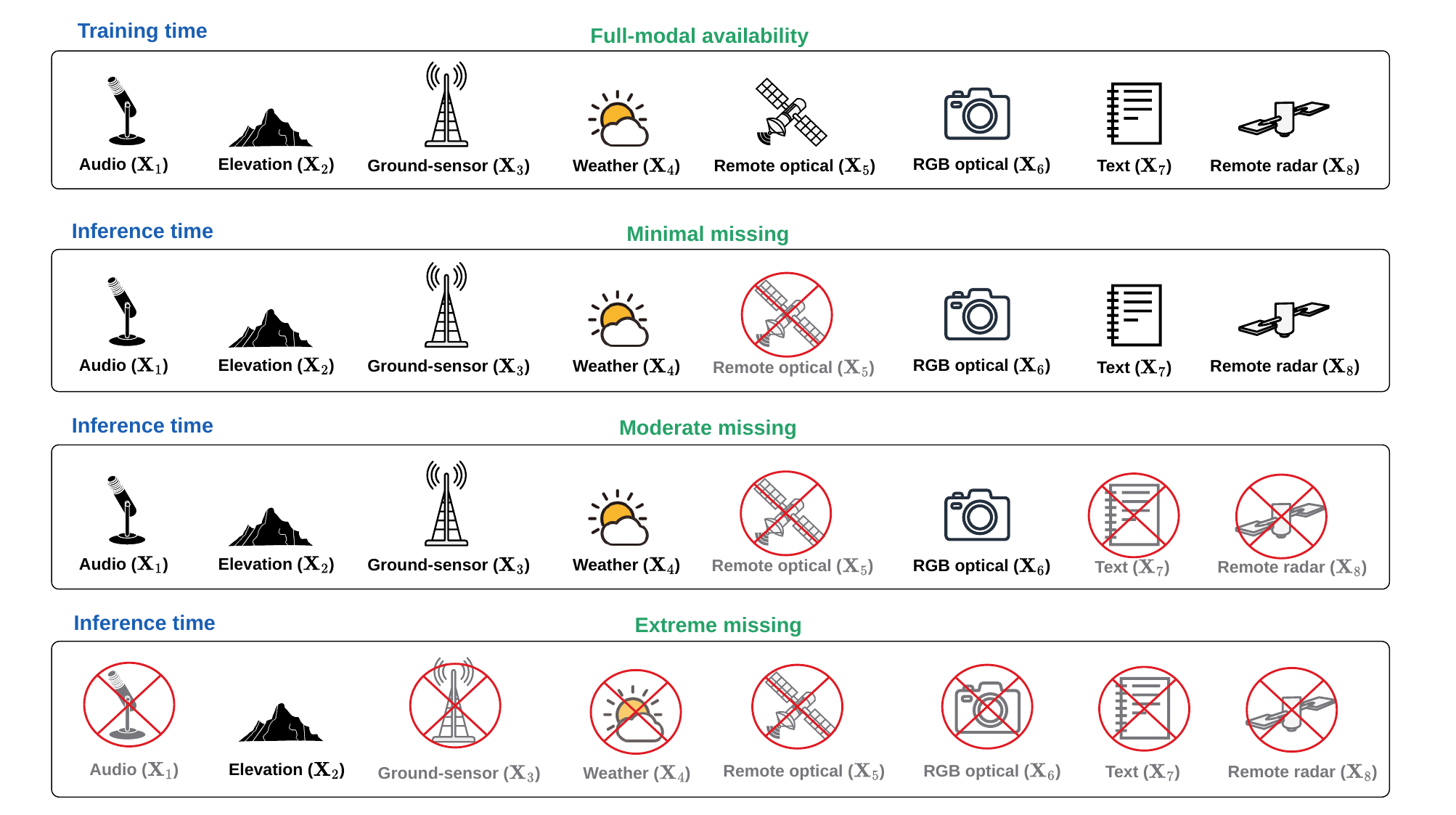}
    \caption{Illustration of three missing modality conditions at inference time: minimal (a missing case when a single modality is missing), moderate (various modalities are missing), and extreme (all-but-one modalities are missing). 
    }
    \label{fig:missing}
\end{figure}

Despite efforts to handle missing modalities, important challenges remain. 
Most existing methods focus on bimodal setups \cite{wu2024deep}, where models are trained with two modalities, and one is missing at inference time.
This limits the applicability of methods such as SMIL \cite{ma2021smil}, DisCoM \cite{ienco2024discom}, MDiCo \cite{mena2025mdico}, and hallucination-based approaches \cite{hoffman2016learning} to broader multi-modal setups.
In addition, some works specifically address the missing of all-but-one modalities, like EmbrNet \cite{choi2019embracenet}, distillation \cite{wu2024deep}, 
and cross-reconstruction-based methods \cite{dong2023simmmdg}.
In contrast, we address the more general scenario of \textbf{missing arbitrary modalities}, where any subset of the training modalities may be available at inference time. 
As illustrated in Fig.~\ref{fig:missing}, this encompasses a broad spectrum of missing modality \textbf{conditions}, from minimal (single modality) to moderate and extreme (all-but-one modalities).

To address missing arbitrary modalities, we formulate a framework grounded in the principles of multi-modal co-learning and knowledge distillation.
Rather than focusing the model design on robust fusion processes as previous works have done \cite{neverova2015moddrop,choi2019embracenet,ma2022multimodal,woo2023towards,dong2023simmmdg,mena2025maug}, we prioritize inter-modal collaboration to encourage each modality to contribute to and benefit from the others.
Concretely, we introduce two decision-level fusion methods that leverage information at both feature- and decision-level to enhance the classification robustness: {\gls{comiss}} and {\gls{fullco}}.
Each method is co-guided to learn modality-shared and modality-specific features with volume-based contrastive loss and a modality discriminant classifier. 
The \gls{comiss} method also adopts \gls{moddrop}, enforcing missing modality predictions to imitate a consensus derived from full-modality data, a strategy we refer to as missing distillation.
In contrast, the \gls{fullco} method employs the mutual distillation strategy to guide individual per-modality predictions toward the full-modality consensus.

We validate our methods via two multi-modal classification benchmarks. The Multi-CropHarvest dataset for crop-type recognition with four sensor modalities, and HL-Opportunity for human activity recognition with nineteen sensor modalities. 
The results demonstrate the classification robustness of our methods against three baselines and seven recent state-of-the-art approaches.
Furthermore, the results reveal complementary strengths between our approaches: the \gls{comiss} method proves particularly effective under minimal missing conditions, while the \gls{fullco} method stands out in moderate to extreme conditions. 
Taken together, these findings highlight the potential of our co-learning strategies for handling missing arbitrary modalities in real-world multi-modal classification scenarios.
\section{Related Work} \label{sec:related}

\subsubsection{Multi-modal Co-learning}
The co-learning paradigm has been applied to multi-modal data to handle noisy modalities and weak supervision \cite{rahate2022multimodal}.
This consists of unimodal models employing a feature-, decision-, or model-based collaboration. 
The feature-based approach usually involves the design of a shared space across modalities. 
For instance, contrastive learning frameworks maximize the cosine-similarity among modalities \cite{yuan2021multimodal}, while Ienco et al. \cite{ienco2024discom} align representations with a modality classifier trained adversarially. 
In decision-based co-learning, the objective is to exchange the unimodal predictive knowledge. 
In image classification, Black et al. \cite{black2024multi} use the mutual distillation strategy to guide per-modality predictions into a full-modality consensus.
In contrast, the model-based approach exchanges components across unimodal networks. For example, Zhang et al. \cite{zhang2024multimodal} propose a modality-shared prediction head with orthogonal gradients among modalities. 
Moreover, Zadeh et al. \cite{zadeh2020foundations} provide theoretical evidence that multi-modal training can improve performance even if a single modality is available at inference time, suggesting the benefit from auxiliary modalities. 

\subsubsection{Missing Modalities}
The literature has addressed the challenge of missing modalities with diverse strategies \cite{wu2024deep}. 
A common technique corresponds to imputing missing data with a zero placeholder. However, this carries strong bias with a drop in predictive performance \cite{ma2022multimodal}.
An alternative consists of designing a robust fusion process in multi-modal learning. Recent models like ShaSpec \cite{wang2023shaspec} and MissModal \cite{lin2023missmodal} use the \gls{moddrop} technique \cite{neverova2015moddrop} to make the multi-modal fusion invariant to missing modalities.
Additionally, missing-modality predictions can be regularized to remain less confident than their full-modality counterparts \cite{ma2023calibrating}, or optimized such that their loss is explicitly higher than that of the full-modality case \cite{li2025simmlm}.
To encourage robustness, Choi et al. \cite{choi2019embracenet} propose to select a random modality (for each feature) in the fusion process. 
Data-driven recovery has also been explored by reconstructing the modalities that are missing from the available ones. Models like SMIL \cite{ma2021smil} and SimMMDG \cite{dong2023simmmdg} reconstruct the data modalities for the underlying full-modality fusion process.
Beyond the focus on robust fusion design, knowledge- and self-distillation frameworks have been employed to increase robustness to missing modalities \cite{mckinzie2023robustness,lin2023missmodal}.
A standard distillation setup consists of a full-modal teacher who guides student models trained with a subset of all training modalities.

\subsubsection{Co-learning for Missing Modalities}
The collaboration of modality-dedicated components has been leveraged to improve robustness when modalities are missing. 
For instance, Hoffman et al. \cite{hoffman2016learning} use a hallucination approach that enforces an unimodal model mimicking the behavior of a potential missing modality. 
Mena et al. \cite{mena2025mdico}, use a feature-based co-learning strategy to enforce unimodal models learning modality-shared and -specific features in a bimodal setup.
Zhang et al. \cite{zhang2024multimodal} use a model-based co-learning strategy that exchanges parameters between unimodal models.
As decision-based co-learning, Mena et al. \cite{mena2025dsensd} introduced mutual distillation in unimodal models tailored for \gls{eo} data classification. 
These works have shown the potential of co-learning for missing modalities.

Among these works, only a limited number address the challenge of missing arbitrary modalities and extend beyond bimodal setups. For instance, EmbrNet \cite{choi2019embracenet}, FCoM \cite{mena2025maug}, and MissModal \cite{lin2023missmodal} 
rely on random modality masking to improve the fusion robustness, yet their performance with missing modalities remains limited.
To this end, we formulate two methods grounded in the principles of multi-modal co-learning and knowledge distillation to enhance the classification robustness in inference scenarios with missing arbitrary modalities.
\section{Methodology} \label{sec:method}

\paragraph{Multi-Modal Setup} \label{sec:method:problem}
Let us consider $\myset{M}$ as the set of $M=|\myset{M}|$ training modalities, and $\myset{X} = \{ \mat{X}_m \}_{m \in \myset{M}}$ the multi-modal input data with label $y\in\{1, \ldots, K\}$. 
At inference time, any arbitrary subset of these modalities $\tilde{\myset{M}} \subseteq \myset{M}$ (with $\tilde{M}=| \tilde{\myset{M}} |$) may be accessible, expressed by $\tilde{\myset{X}}= \{ \mat{X}_m \}_{m \in \tilde{\myset{M}}}$.  
We define this scenario, encompassing any missing modality case, as \textbf{missing arbitrary modalities}.

To address this challenge, we introduce two multi-modal methods employing decision-level fusion, also known as late fusion. 
In concrete, we employ a simple fusion process where per-modality probability predictions $\hat{\myvec{y}}_m \in [0,1]^K$ are averaged to yield the full-modality estimation (called \textbf{consensus}), given by $\hat{\myvec{y}}_{\text{full}}  = M^{-1} \cdot \sum_{m\in \myset{M}} \hat{\myvec{y}}_m$.
Thus, if modalities are missing at inference time, they are just disregarded from the aggregation, expressed by $\tilde{\myvec{y}} = \tilde{M}^{-1} \cdot \sum_{m \in \tilde{\myset{M}}} \hat{\myvec{y}}_m $.

For optimization, we use a standard cross-entropy loss to guide the main prediction with full-modality data, defined as
\begin{equation}
    \mathcal{L}_{\text{main}} = \mathcal{L}_{\text{CE}} \left(y, \hat{\myvec{y}}_{\text{full}} \right) \ ,
\end{equation}
where $\letterfunc{L}_{\text{CE}}(p,\myvec{q}) = - \sum_k \mathbbm{1}(p=k) \cdot \log{q_k}$ is the loss function between the true label $p$ and probabilities $\myvec{q}$, and $\mathbbm{1}(\cdot)$ is the indicator function.
We introduce additional loss terms at the feature-level (via co-learning) and at the decision-level (via knowledge distillation)  as follows.

\subsection{Feature-level Learning Criteria} \label{sec:method:feature}

We assume that training modalities have shared (invariant among modalities) and specific (unique to each modality) information among them relevant for the classification. 
The \emph{shared space} corresponds to class-relevant features that can be extracted from either modality, while the \emph{specific space} contains class-relevant features that can only be extracted from a specific modality. 
For instance, a high-resolution and a low-resolution optical satellite image have common data (the optical part) and specific data (related to the differences in spatial resolutions).
In concrete, we use modality-dedicated encoders that learn to extract both the shared $\myvec{z}_{m}^{\text{sha}} \in \myset{R}^d$ and specific $\myvec{z}_{m}^{\text{spe}} \in \myset{R}^d$ features explicitly, given by 
\begin{equation} \label{eq:encoder}
    \myvec{z}_{m}^{\text{sha}}, \myvec{z}_{m}^{\text{spe}} = \letterfunc{E}_m(\mat{X}_m) \ .
\end{equation}

For learning the specific features per modality, we use a modality discriminant loss function (based on the cross-entropy), expressed by 
\begin{equation}
    \mathcal{L}_{\text{mod}} = \frac{1}{M} \sum_{m\in \myset{M}} \mathcal{L}_{\text{CE}} \left( m , \letterfunc{P}^{\text{spe}} \left( \myvec{z}_m^{\text{spe}} \right) \right) \ ,
\end{equation}
where $\letterfunc{P}^{\text{spe}}(\cdot)$ is an auxiliary linear classifier that is fed with the specific features from either modality and has to predict the correct one.
In this way, the specific features per sample have to be linearly distinguished among modalities.

For learning the shared features among modalities, we use a volume-based contrastive loss function \cite{cicchettigramian}, defined by
\begin{equation}
    \mathcal{L}_{\text{cont}} =  \frac{1}{2} \left( \mathcal{L}_{\text{GRAM}} \left( \myvec{z}_a^{\text{sha}} , \myset{Z}_a^{\text{sha}}; \gamma \right) + \mathcal{L}_{\text{GRAM}} \left( \myset{Z}_a^{\text{sha}}, \myvec{z}_a^{\text{sha}} ); \gamma \right)  \right) \ ,
\end{equation}
where $\mathcal{L}_{\text{GRAM}}(\cdot, \cdot ; \gamma)$ is the GRAM multi-modal contrastive loss defined in \cite{cicchettigramian} and parametrized by a scalar $\gamma$, $a \in \myset{M}$ is the anchor modality, and $\myset{Z}_a^{\text{sha}} = \{ \myvec{z}_m^{\text{sha}} : m \in \myset{M}/a \}$. We use $\gamma = 0.07$, following \cite{mena2025mdico}.
The minimization of the Gramian volume among the shared features is applied over all modalities for each sample, enforcing them to be closer in space.

\subsubsection{Individual Prediction} To obtain per-modality predictions, our model concatenates the modality-shared and modality-specific features (from Eq.~\eqref{eq:encoder}) followed by a modality-dedicated linear head $\letterfunc{P}_m(\cdot)$. This is given by $\hat{\myvec{y}}_m =  \letterfunc{P}_m(\myvec{z}_{m}^{\text{sha}} || \myvec{z}_{m}^{\text{spe}})$. 

\subsection{Decision-level Learning Criteria} \label{sec:method:decision}
We consider two knowledge distillation variants used at the decision-level to improve the predictive robustness to missing arbitrary modalities.

\begin{figure}[!t]
  \centering
  \includegraphics[width=1.01\linewidth]{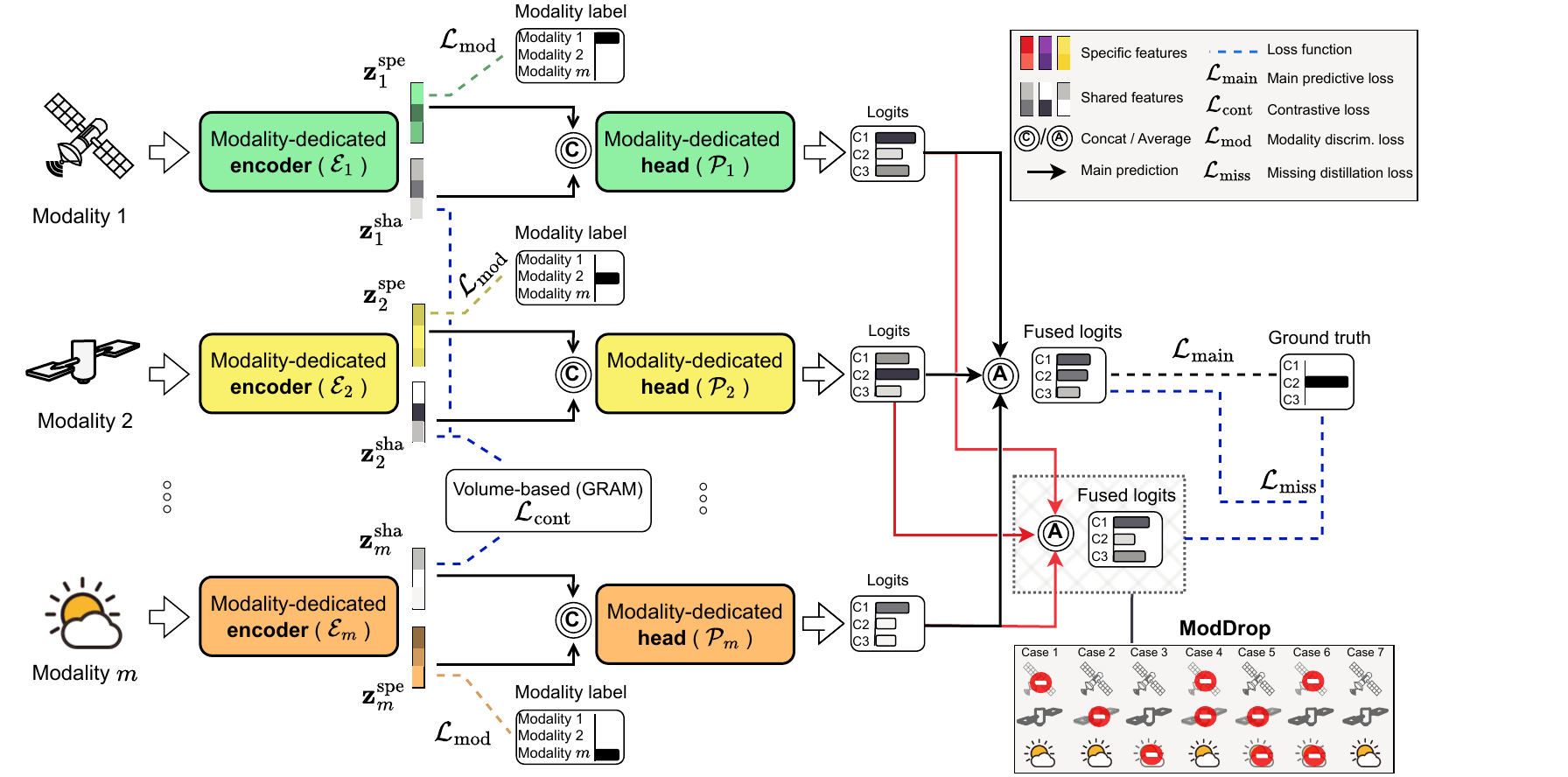}
  \caption{Illustration of the \gls{comiss} method with four loss functions.} \label{fig:comiss}
\end{figure}
\subsubsection{Missing Distillation} 
Expecting any missing modality condition (see Fig.~\ref{fig:missing}), we incorporate the \gls{moddrop} technique at the decision-level to expose the model to missing modalities during training. 
This missing modality prediction is expressed by $\hat{{\vec{y}}}_{\text{miss}} = M_{\text{drop}}^{-1} \cdot \sum_{m \in \myset{M}} (1-d_m) \cdot \hat{\myvec{y}}_m $, where $d_m \sim \text{Bern}(\alpha)$ the randomly drawn decision if modality $m$ is masked out, and $M_{\text{drop}} = \sum_{m \in \myset{M}} d_m$ the number of modalities available.
However, instead of drawing a single random case per sample, we exhaustively simulate all missing modality cases, i.e. $2^{M} -1$ combinations.
Then, we enforce this missing modality prediction $\hat{\myvec{y}}_{\text{miss}}$ to imitate the consensus $\hat{\myvec{y}}_{\text{full}}$, as well as the ground truth $y$, by the following loss function 
\begin{equation}
    \mathcal{L}_{\text{miss}} =  \mathcal{L}_{\text{CE}} \left( y , \hat{\myvec{y}}_{\text{miss}} \right) + \lambda \cdot \mathcal{L}_{\text{KD}}\left(\hat{\myvec{y}}_{\text{full}}, \hat{\myvec{y}}_{\text{miss}} ; \tau \right) \ ,
\end{equation}
where $\lambda$ is a weighting factor (we use $\lambda=\tau^2$ following \cite{mena2025dsensd}), and $\mathcal{L}_{\text{KD}}(\cdot, \cdot ; \tau)$ is the knowledge distillation function parametrized by the temperature $\tau$ (we use $\tau=0.5$ following \cite{mena2025dsensd}). We refer to this strategy as \textbf{missing distillation}.
This \textbf{\acrfull{comiss}} method is optimized via an unweighted sum of all afore-mentioned loss terms, expressed by 
\begin{equation} \label{eq:total:comiss}
    \mathcal{L}_{\text{total}} = \mathcal{L}_{\text{main}} + \mathcal{L}_{\text{mod}} + \mathcal{L}_{\text{cont}} + \mathcal{L}_{\text{miss}} \ .
\end{equation}
This method, illustrated in Fig.~\ref{fig:comiss}, follows a feature-based co-learning (in $ \mathcal{L}_{\text{mod}}$ and $\mathcal{L}_{\text{cont}}$) combined with a distillation process (in $\mathcal{L}_{\text{miss}}$).

\begin{figure}[!t]
  \centering
  \includegraphics[width=1.1\textwidth]{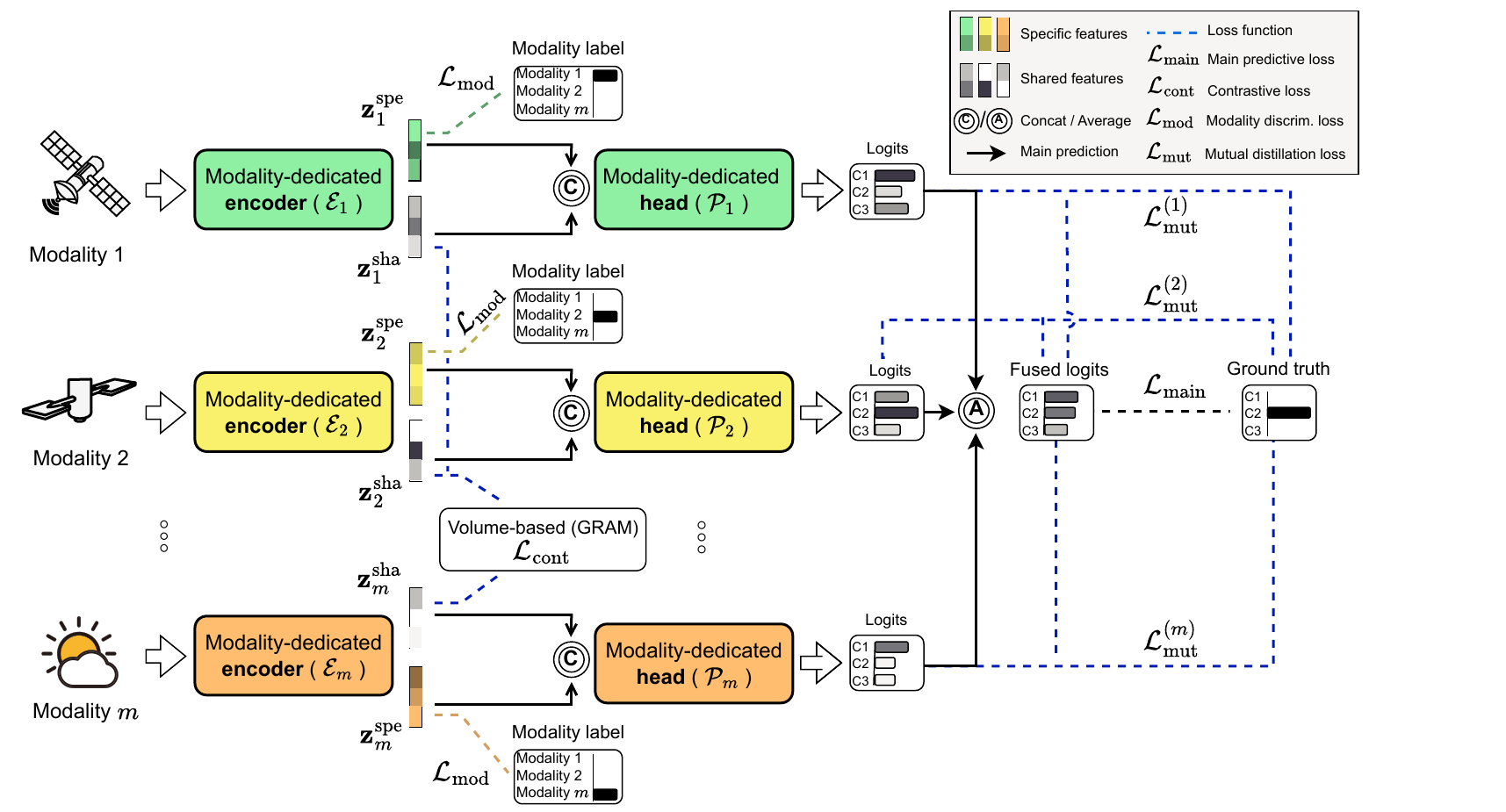}
  \caption{Illustration of the \gls{fullco} method with four loss functions.} \label{fig:fullco}
\end{figure}
\subsubsection{Mutual Distillation} 
To address extreme missing conditions (See Fig.~\ref{fig:missing}), we follow a decision-based co-learning strategy called mutual distillation, where each per-modality prediction $\hat{\myvec{y}}_m$ has to imitate both the consensus $\hat{\myvec{y}}_{\text{full}}$ and ground truth ${y}$.
The loss function of this mutual distillation strategy considers the average of the individual per-modality distillations, given by
\begin{align}
    \mathcal{L}_{\text{mut}} &= \frac{1}{M} \sum_{m\in \myset{M}} \mathcal{L}_{\text{mut}}^{(m)} \\
    \mathcal{L}_{\text{mut}}^{(m)} &= \mathcal{L}_{\text{CE}} \left( y , \hat{\myvec{y}}_m \right) + \lambda \cdot \mathcal{L}_{\text{KD}} \left( \hat{\myvec{y}}_{\text{full}}, \hat{\myvec{y}}_m ; \tau \right) \ ,
\end{align}
where $\lambda=\tau^2$ and $\tau=0.5$, following the same criteria as for the \gls{comiss} method.
This \textbf{\acrfull{fullco}} method is learned by optimizing an unweighted sum of the following loss terms:
\begin{equation} \label{eq:total:fullco}
    \mathcal{L}_{\text{total}} = \mathcal{L}_{\text{main}} + \mathcal{L}_{\text{mod}} + \mathcal{L}_{\text{cont}} + \mathcal{L}_{\text{mut}}\ .
\end{equation}
This method, illustrated in Fig.~\ref{fig:fullco}, follows a full co-learning strategy, with a feature-based (in $ \mathcal{L}_{\text{mod}}$ and $\mathcal{L}_{\text{cont}}$) and decision-based (in $\mathcal{L}_{\text{mut}}$) collaboration.

Our methods differ in how the decision-level learning is carried out: through the missing distillation ($\mathcal{L}_{\text{miss}}$) in \gls{comiss} or mutual distillation ($\mathcal{L}_{\text{mut}}$) in \gls{fullco}.
\section{Experiments} \label{sec:exp}


\subsection{Experimental Setup}

\subsubsection{Datasets} 
We use the following multi-modal classification benchmarks. 

\paragraph{Multi-CropHarvest}\footnote{\url{https://github.com/nasaharvest/cropharvest} (Accessed 27.07.2026).} We consider a crop-type recognition problem by using the CropHarvest dataset. 
This benchmark contains 29\,642 samples around the globe between 2016 and 2021.
The label covers ten different crop-type groups (\textit{other, beverage spice, cereals, leguminous, fruits nuts, root tuber, oilseeds, vegetables melons, sugar, non crop}).
The input data consist of three multi-temporal sensor modalities, at $10 [m]$ spatial resolution: multi-spectral optical data from Sentinel-2, radar data from Sentinel-1, and weather variables. In addition, topographic features (mono-temporal modality) are available for each sample.
Following previous research \cite{mena2025dsensd,mena2025mdico}, we perform a 10-fold cross-validation.

\paragraph{HL-Opportunity}\footnote{\url{www.opportunity-project.eu/challengeDataset.html} (Accessed 27.07.2026).} We consider a high-level activity recognition problem by using the Opportunity dataset. 
This benchmark consists of data from six subjects performing different activities (five high-level: \textit{relaxing, coffee time, early morning, cleanup, sandwich time}).
There are 19 modalities, corresponding to various sensors positioned on the body of the subjects (e.g. food, back, arms, hip).
Following the evaluation setup by Choi et al. \cite{choi2019embracenet}, a total of 264\,279, 6\,528, and 12\,688 samples are available for training, validation, and testing, respectively.

\subsubsection{Comparison}
We compare our methods against the following approaches.

\paragraph{Baselines} We consider unimodal models trained exclusively for each modality (i.e. no multi-modal interaction during training). As multi-modal baselines, we include two models employing feature- (Feat) and decision-level (Dec) fusion, respectively, where missing modalities are imputed with zeros at inference time.

\paragraph{Competitors} We include seven methods from the recent literature, general enough to accommodate multi-modal beyond bimodal setups.
Five methods use feature-level fusion: EmbrNet \cite{choi2019embracenet}, a method that randomly selects a single modality per feature, and the following \gls{moddrop}-based approaches: MissModal \cite{lin2023missmodal}, ShaSpec \cite{wang2023shaspec}, FCoM \cite{mena2025maug}, and SimMLM \cite{li2025simmlm}. 
In addition, two decision-level fusion methods that leverage the mutual distillation strategy: DML \cite{zhang2018deep}, and DSensD$^+$ \cite{mena2025dsensd}.

\subsubsection{Evaluation details}
For the classification assessment, we calculate the \gls{f1} score with full-modality data and under every missing modality case.
For each missing modality condition (i.e. minimal, moderate, and extreme), we average the \gls{f1} values of the corresponding missing modality cases. 
We also consider the \textbf{macro missing} performance as the average between the following settings: full-modality data, minimal missing, moderate missing, and extreme missing conditions. 
In addition, the method rank (according to performance, i..e rank 1= best) is calculated and averaged across all missing modality cases.

\paragraph{Implementation} We apply a z-score normalization on the input data. 
We use standard encoder architectures for the selected datasets \cite{mena2025dsensd,choi2019embracenet}. This corresponds to a 1D CNN for multi-temporal modalities, and an MLP for mono-temporal ones. For each encoder, we use two layers with 128 units and  20\% of dropout on all encoders in the Multi-CropHarvest dataset, and four layers with 64 units for the HL-Opportunity.
We apply a layer normalization to the per-modality predictions to scale the logits magnitudes.
For optimization, we use AdamW 
with a learning rate of $10^{-3}$, batch-size of 128, and early stopping. The stopping considers the \gls{f1} score (macro missing) calculated on the validation set.
We use weights in the main predictive loss function $\mathcal{L}_{\text{main}}$ (inverse to the number of samples per class) to cope with class imbalance.
For all competitors, we retain the default hyperparameter settings as reported in their original works.

\subsection{Experimental Results}

\begin{table}[!b]
    \centering
    \caption{\acrshort{f1} score in the crop-type recognition task ({Multi-CropHarvest}). The sensor modalities are optical (O), radar (R), weather (W), and topographic (T). \textbf{Bold} indicates the best mean performance. \underline{Underlined} indicates values that are not statistically different from the best method (Welch's t-test, $p>0.05$).
    \footnotesize{$^*$Averaged over available cases}.}  \label{tab:main:cropm}
    \begin{tabularx}{\textwidth}{ccccCccCCCCCcCCC} 
         \toprule
         O & R  & W & T & Uni-modal & Feat & Dec & Embr-Net & Miss-Modal & Sha-Spec & FCoM & Sim-MLM &  DML &  DSen-sD$^+$ &  \gls{comiss} & Full-Co \\
         \midrule
         $\checkmark$ & $\checkmark$ & $\checkmark$ & $\checkmark$& $71.1$ &	$73.3$ &	$73.6$ &	$72.1$ &	$70.6$ &	$67.1$ &	$75.1$ &	$73.9$ &	$66.5$ &	$74.3$ &		$\best{79.9}$ &	$\secondbest{79.7}$   \\
         \midrule
         $\times$ & $\checkmark$ & $\checkmark$ & $\checkmark$ &  & $57.5$ &	$56.5$ &	$56.6$ &	$63.7$ &	$56.4$ &	$64.4$ &	$62.3$ &	$58.7$ &	$65.1$  &	$\best{67.8}$ &	$\secondbest{67.7}$ \\
         $\checkmark$ & $\times$ & $\checkmark$ & $\checkmark$ & & $66.1$ &	$69.4$ &	$69.6$ &	$68.1$ &	$64.9$ &	$73.2$ &	$71.7$ & $64.2$ &	$73.1$  &	$\best{79.0}$ &	$\secondbest{78.9}$   \\
         $\checkmark$ & $\checkmark$ & $\times$ & $\checkmark$ & & $65.0$ &	$59.7$ &	$65.6$ &	$69.5$ &	$65.9$ &	$74.8$ &	$73.2$ &	$69.0$ &	$74.5$ &	$\secondbest{79.3}$ &	$\best{79.4}$ \\
         $\checkmark$ & $\checkmark$ & $\checkmark$ & $\times$ & & $70.3$ &	$73.0$ &	$72.5$ &	$70.5$ &	$67.0$ &	$75.2$ &	$73.8$ &	$67.2$ &	$74.7$  &	$\best{79.9}$ &	$\secondbest{79.7}$  \\
         \cmidrule{6-16}
         \multicolumn{4}{c}{minimal} &  & $64.7$ &	$64.7$ &	$66.1$ &	$67.9$ &	$63.6$ &	$71.9$ &	$70.3$ &	$64.8$ &	$71.9$ &	$\best{76.5}$ &	$\secondbest{76.4}$  \\
         \midrule
         \multicolumn{4}{c}{moderate} &  & $51.8$ &	$51.4$ &	$56.2$ &	$62.0$ &	$55.6$ &	$65.1$ &	$62.5$ &	$59.5$ & $65.7$ &	$\secondbest{69.3}$ &	$\best{69.6}$  \\
         \midrule
         $\checkmark$ & $\times$ & $\times$ & $\times$ & $72.5$ &	$57.8$ &	$56.4$ &	$64.4$ &	$67.4$ &	$63.4$ &	$73.3$ &	$70.8$ &	$77.8$ &	$73.6$ &	$\secondbest{78.5}$ &	$\best{78.8}$ \\
          $\times$& $\checkmark$ & $\times$ & $\times$ & $55.6$ &	$37.1$ &	$31.8$ &	$38.8$ &	$54.2$ &	$43.5$ &	$54.2$ &	$51.0$ &	$56.6$ &	$56.9$ &	${57.9}$ &	$\best{59.2}$   \\
           $\times$& $\times$ & $\checkmark$ & $\times$ & $46.5$ &	$30.9$ &	$34.8$ &	$39.9$ &	$47.2$ &	$39.7$ &	$46.1$ &	$42.7$ &	$47.4$ &	$48.2$ &	$\secondbest{51.3}$ &	$\best{52.0}$ \\
          $\times$& $\times$ & $\times$ & $\checkmark$ & $21.8$ &	$10.0$ &	$3.1$ &	$8.3$ &	$22.9$ &	$12.9$ &	$21.5$ &	$18.3$ &	$23.9$ &	$28.4$ &	${31.2}$ &	$\best{33.5}$  \\
          \cmidrule{5-16}
          \multicolumn{4}{c}{extreme} & $49.1$ &	$34.0$ &	$31.5$ &	$37.8$ &	$47.9$ &	$39.9$ &	$48.8$ &	$45.7$ &	$49.8$ &	$\secondbest{51.8}$ &		$\secondbest{54.7}$ &	$\best{55.8}$   \\
         \midrule 
          \multicolumn{4}{c}{Macro} & $60.1^*$ &	$55.8$ &	$55.4$ &	$58.0$ &	$62.2$ &	$56.3$ &	$65.4$ &	$63.2$ &	$61.2$ & $65.8$  &	$\secondbest{70.1}$ &	$\best{70.4}$ \\
          \midrule
          \multicolumn{4}{c}{Ranking} & & $9.7$ &	$9.6$ &	$8.0$ &	$6.5$ &	$9.2$ &	$3.9$ &	$5.8$ &	$6.8$ &	${3.5}$ &	$\best{1.5}$ &	$\best{1.5}$ \\
         \bottomrule
    \end{tabularx}
\end{table}
We report the crop-type recognition results in Table~\ref{tab:main:cropm}. 
Here, the lowest drop in performance occurs under minimal missing conditions, varying whether the optical, radar, weather, or topographic modality is missing. 
In the Dec baseline, this ranges from $0.6$ points when the topographic is missing to $17.1$ points when the optical is missing.
In contrast, the \gls{comiss} method reduces these drops to $0$ and $12.1$ points, respectively.
Overall, the drop in performance is more evident in the extreme missing conditions. 
In the baselines, there is around $40$ points of performance drop in the extreme (average) condition. 
In this case, our \gls{comiss} method has a drop of $25.2$, while \gls{fullco} has $23.9$ points.
In addition, our methods outperform all competitors with significant differences in each missing condition, due to different advantages.
The \gls{comiss} method outperforms all approaches in full-modality data, as well as in minimal missing conditions, except when missing the weather modality.
The \gls{fullco} method achieves the best results under moderate and extreme missing conditions. 
This is expected, as the \gls{fullco} method is explicitly trained to handle unimodal data (via the distillation of per-modality predictions).
Besides, \gls{fullco} achieves the second-best results in full-modality data and minimal missing conditions.
These complementary advantages make our methods tie as the best average ranking with a $1.5$ rank.

Finally, the macro missing results in the Multi-CropHarvest dataset indicate that, regardless of the modalities missing at inference time, the expected \gls{f1} performance is $70.1$ and $70.4$ for \gls{comiss} and \gls{fullco} methods, respectively. 
This value improves around 4 points compared to the closest competitor (DSensD$^+$), and up to 15 points compared to the Feat and Dec baselines. 
Moreover, the improvements achieved by our methods are statistically significant in all cases. 
Besides, no significant differences are observed among our methods across the inference scenarios, except when radar or topographic modalities are available.

\begin{table}[!t]
    \centering
    \caption{\acrshort{f1} score in the high-level activity recognition task ({HL-Opportunity}). \textbf{Bold} indicates the best mean performance. \underline{Underlined} indicates values that are not statistically different from the best method (Welch's t-test, $p>0.05$).
    \footnotesize{$^*$Averaged over available cases}
    }  \label{tab:main:hl}
    \begin{tabularx}{\textwidth}{cCccCCCCCcCCC} 
         \toprule
         Inference & Uni-modal & Feat & Dec & Embr-Net & Miss-Modal & Sha-Spec  & FCoM & Sim-MLM &  DML & DSen-sD$^+$  & \gls{comiss} & Full-Co \\
         \midrule
         full-modal & $77.4$ & $71.9$ &	$69.6$ &	$71.0$ &	$\secondbest{78.7}$ &	$77.0$ &	$\secondbest{78.8}$ &	$\secondbest{78.2}$ &	$76.0$ &	$\best{79.5}$ &	$78.1$ &	$\secondbest{79.0}$ \\
        minimal &  & $70.6$ &	$68.7$ &	$70.7$ &	$78.5$ &	$76.6$ &	$78.5$ &	$77.9$ &	$76.0$ &	$\best{79.4}$&	$77.7$ &	${78.7}$  \\
         moderate & & $57.5$ &	$56.4$ &	$62.2$ &	$72.2$ &	$67.6$ &	$69.9$ &	$69.7$ &	$73.5$ &	$\best{74.9}$ &	$70.0$ &	$\secondbest{74.0}$  \\
          extreme  & $59.1$  &$28.0$ &	$27.1$ &	$35.1$ &	$49.6$ &	$32.3$ &	$41.1$ &	$42.6$ &	$\best{60.5}$ &	${57.0}$ &	$43.3$ &	$56.4$   \\
         \midrule
         Macro & $68.2^*$ & $57.0$ &	$55.5$ &	$59.7$ &	$69.8$ &	$63.4$ &	$67.0$ &	$67.1$ &	$\secondbest{71.5}$ &	$\best{72.7}$ & $67.0$ &	$\secondbest{72.0}$ \\
         \midrule
         Ranking & & $10.1$ &	$10.6$ &	$8.9$ &	$3.7$ &	$7.5$ &	$5.4$ &	$5.6$ &	$4.3$ &	$\best{1.4}$ &	$6.1$ &	${2.4}$   \\
         \bottomrule
    \end{tabularx}
\end{table}
The results in the activity recognition task are displayed in Table~\ref{tab:main:hl}. We only report the average of the three missing conditions, as with 19 sensor modalities, there are $2^{19}-1$ missing modality cases possible.
Similarly to previous results, the performance degrades the least under minimal missing conditions and more noticeably under extreme missing ones.
The potential information redundancy among the 19 sensor modalities explains the limited performance drop when a single modality is missing (i.e., 18 modalities are still available).
In our methods, the performance decreases by just 0.3 points with \gls{fullco} and 0.4 with \gls{comiss}. 
Overall, the best results are achieved by the DSensD$^+$ method in all missing conditions except for the extreme missing one, where DML obtains the best results. 
Among our methods, \gls{fullco} competes with DSensD$^+$ and DML methods, obtaining the second-best results across all missing conditions (except extreme), and being statistically similar to the best results in both full-modal and moderate conditions.
Moreover, the average ranking shows \gls{fullco} as the second-best method overall with a $2.4$ rank.
At last, the macro missing results indicate that, regardless of the modalities missing at inference time, the expected \gls{f1} performance of the \gls{fullco} method remains $72.0$. This represents a marginal difference of only $0.7$ points compared to DSensD+, which is not statistically significant.

\begin{figure}[t!]
    \centering
    \subfloat
    {\includegraphics[width=0.75\linewidth]{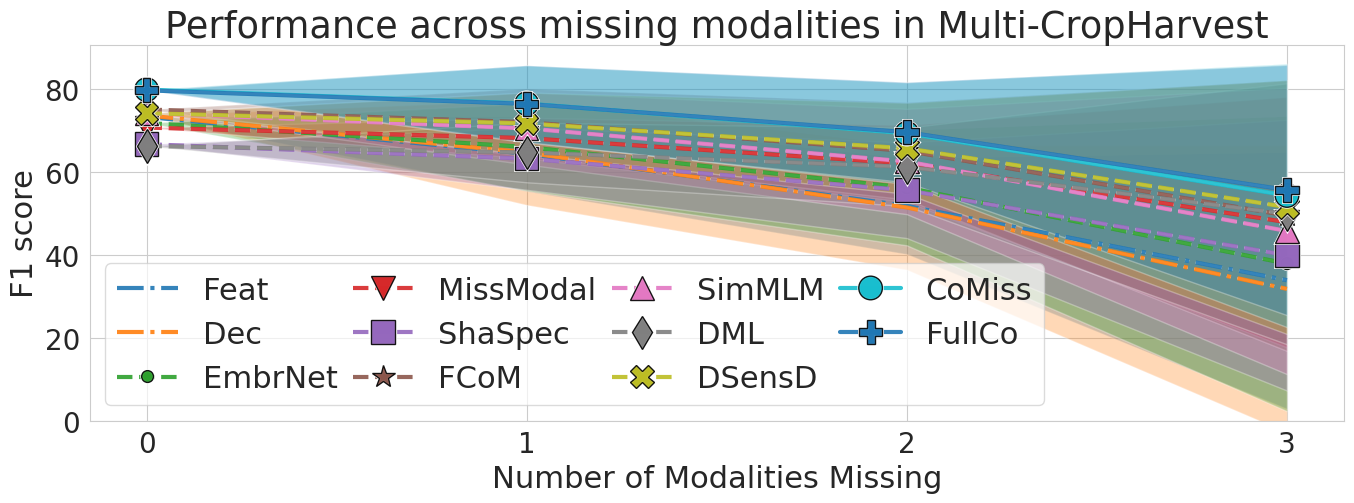}}\\
    \vspace{-0.1cm}
    \subfloat
    {\includegraphics[width=0.75\linewidth]{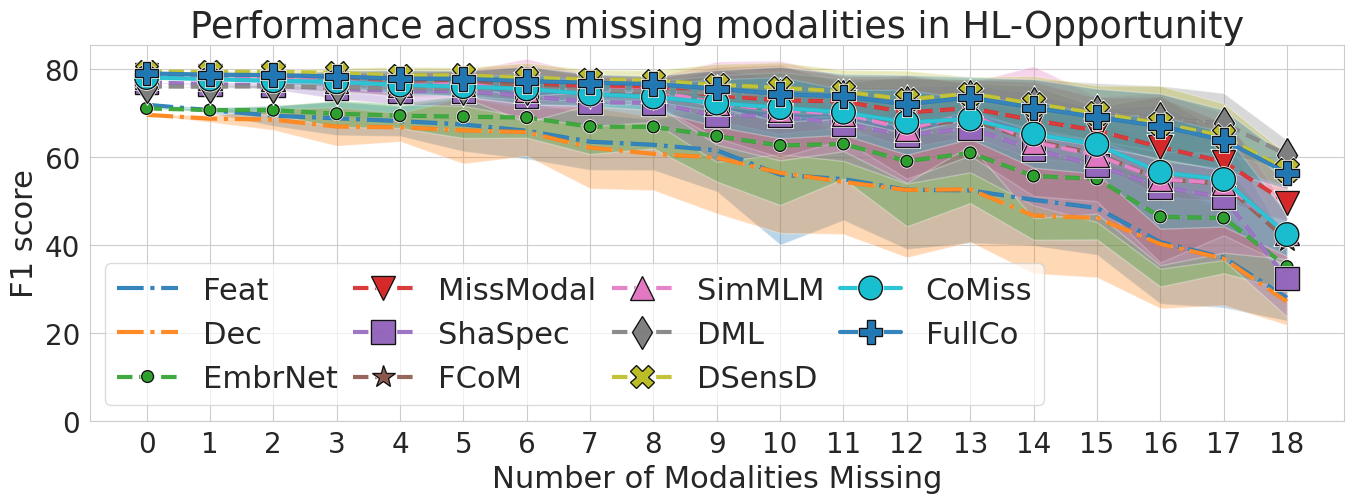}}
    \vspace{-0.1cm}
    \caption{\gls{f1} score by increasing the number of modalities missing at inference time. The 95\% confidence interval is included as shaded colors for each method.} 
    \label{fig:increase:miss}
\end{figure}
\paragraph{Increasing missing modalities}
Figure~\ref{fig:increase:miss} shows that the classification performance decreases non-linearly as the number of missing modalities increases.
This drop is more pronounced for the baseline methods, which lack any dedicated mechanism to handle missing modalities. 
The class-relevant redundancy among the 19 modalities is noted in the HL-Opportunity, where the performance remains mostly unchanged until five modalities are missing.
Moreover, the performance difference and variability between the methods increase considerably as more modalities are missing. 
This trend highlights the growing robustness challenge as we move from minimal to moderate and extreme missing conditions.
Similar to previous results, the performance relates to the considered dataset. In Multi-CropHarvest, \gls{fullco} and \gls{comiss} methods achieve the best robustness curves along missing modalities, while in HL-Opportunity, the \gls{fullco} has comparable behavior to the best competing approaches (i.e. DSensD$^+$ and DML).

\subsection{Ablation Results}

\begin{table}[!t]
    \centering
    \caption{\acrshort{f1} score in the {Multi-CropHarvest} dataset under different configurations in our methods. The \textbf{best} and \underline{second-best} values are highlighted.}  \label{tab:abl}
    \begin{tabularx}{\textwidth}{lCCCCC} 
         \toprule
         Ablation case & Full-modal & Minimal & Moderate & Extreme & Macro \\
         \midrule
         \gls{comiss} & $\best{79.9}$ & $\best{76.5}$ & $\best{69.3}$ & $\secondbest{54.7}$ & $\best{70.1}$  \\
         $\;$ w/o $\mathcal{L}_{\text{mod}}$ & $\best{79.9}$ & $\best{76.5}$ & $\best{69.3}$ & $54.4$ & $\secondbest{70.0}$ \\
         $\;$ w/o $\mathcal{L}_{\text{cont}}$ & $79.2$ & $75.7$ & $\secondbest{68.5}$  & $54.0$ & $69.4$ \\
         $\;$ w/o $\mathcal{L}_{\text{mod}}$ and $\mathcal{L}_{\text{cont}}$ & $79.2$ & $75.6$ & $\secondbest{68.5}$ & $54.1$ & $69.3$ \\
         $\;$ pair-wise contrastive & $77.5$ & $75.4$ & $\secondbest{68.5}$ & $53.4$ & $68.7$ \\
         \midrule
         \gls{fullco} & $\best{79.7}$ & $\best{76.4}$ & $\best{69.6}$ & $\best{55.8}$ & $\best{70.4}$  \\
         $\;$ w/o $\mathcal{L}_{\text{mod}}$ & $\secondbest{79.5}$ & $\secondbest{76.3}$ & $\secondbest{69.5}$ & $\secondbest{55.5}$ & $\secondbest{70.2}$ \\
         $\;$ w/o $\mathcal{L}_{\text{cont}}$ & $78.8$ & $75.5$ & $68.7$ & $55.0$ & $69.5$ \\
         $\;$ w/o $\mathcal{L}_{\text{mod}}$ and $\mathcal{L}_{\text{cont}}$ & $78.9$ & $75.7$ & $68.9$ & $55.2$ & $69.7$ \\
         $\;$ pair-wise contrastive & $77.4$ & $74.1$ & $67.5$ & $54.0$ & $68.2$ \\
         \midrule
          $\;$ w/o $\mathcal{L}_{\text{mut}}$ or $\mathcal{L}_{\text{miss}}$ & $79.8$ & $76.1$ & $68.7$ & $53.4$ & $69.5$ \\
         \bottomrule
    \end{tabularx}
\end{table}
In this subsection, we vary individual components to isolate the key factors characterizing the behavior of our methods. These results are reported in Table~\ref{tab:abl} for the Multi-CropHarvest dataset, while similar results are observed in HL-Opportunity.
We observe that the most relevant loss term is the contrastive one ($\mathcal{L}_{\text{cont}}$), followed by the decision-level term, either mutual ($\mathcal{L}_{\text{mut}}$) or missing ($\mathcal{L}_{\text{miss}}$) distillation.
On the other hand, the loss term that least affects performance is the modality discriminant one. 
This indicates that for our co-learning method, the modality-shared information is more crucial than the modality-specific counterpart.
Furthermore, we compare the volume-based contrastive loss against its standard pair-wise version (computed over all pairs \cite{yuan2021multimodal}). These results suggest that the volume-based formulation is more effective for missing conditions, yielding a more structured modality-shared space within our approach.

\begin{figure}[t!]
    \centering
    \includegraphics[width=0.75\linewidth]{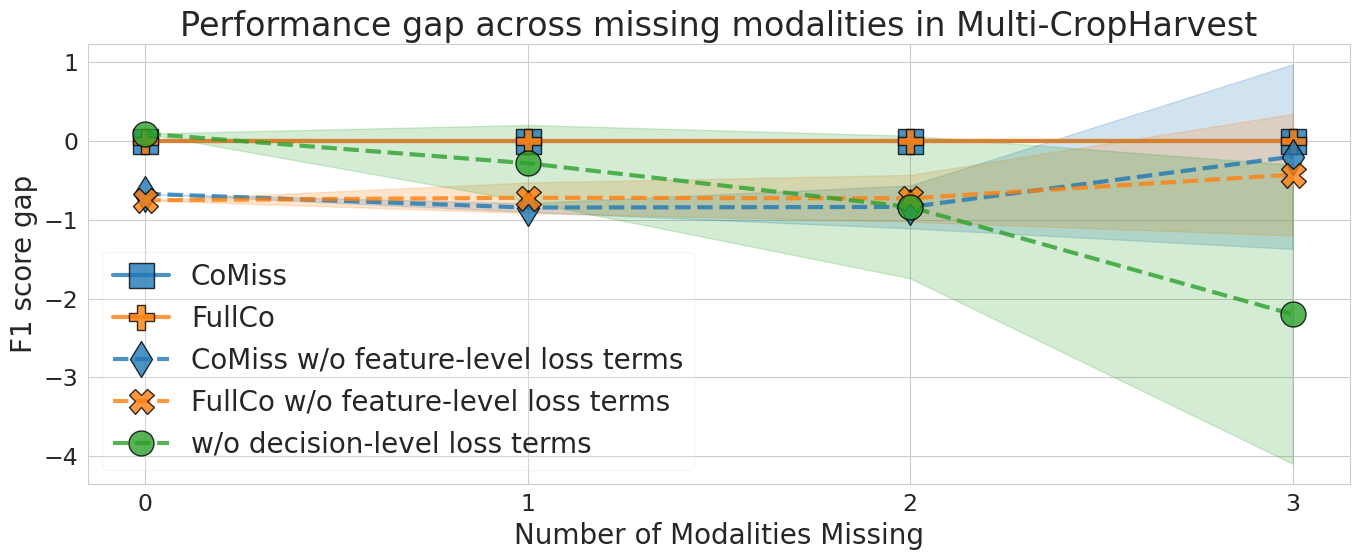}
    \vspace{-0.1cm}
    \caption{\gls{f1} gap between the default version of our methods and different variants.}
    \label{fig:increase:miss:abl}
\end{figure}
We display the performance gap (difference between the \gls{f1} score of a variant and the default version of our approaches) in Fig.~\ref{fig:increase:miss:abl}. 
In this case, removing the feature-level loss terms ($\mathcal{L}_{\text{mod}}$ and $\mathcal{L}_{\text{cont}}$) causes the largest performance drop when few modalities are missing. 
Conversely, removing the decision-level loss term (either $\mathcal{L}_{\text{mut}}$ or $\mathcal{L}_{\text{miss}}$) degrades performance most when more modalities are missing.
This reflects the complementarity of our feature- and decision-level learning criteria: feature-level collaboration is key to robustness when facing moderate to minimal missing conditions, while decision-level distillation is crucial when facing moderate to extreme missing conditions at inference time.

 \subsection{Model Comparison \& Limitations}

The \gls{comiss} method has a scalability limitation that future work should address, as its training complexity scales as $\mathcal{O}(2^M)$. For example, with $M=10$ modalities, it must simulate 1023 missing-modality combinations. 
Given this limitation and its weaker performance under severe missing conditions, we recommend \gls{fullco}, particularly for moderate to extreme missing inference scenarios. 
When a few modalities are available for training (i.e., $M<10$), \gls{comiss} remains a suitable choice, especially for full-modality and minimal missing inference scenarios.

Both of our methods share similarities to MDiCo \cite{mena2025mdico}. Despite the learning of a modality-shared space by contrastive learning (Sec.~\ref{sec:method:feature}), we use a volume-based version instead of the pairwise loss as in \cite{mena2025mdico}. 
However, MDiCo is a purely multi-modal co-learning approach, i.e. no fusion is performed among modalities.
Thus, MDiCo does not employ any decision-level learning criteria as us (Sec.~\ref{sec:method:decision}).

We validate our approaches in two applications, while conventional multimodal benchmarks (e.g., image-text-audio) can demonstrate their generalization beyond sensor-based domains.
Future work should also consider adaptive modeling for varying amounts of modality-shared and modality-specific information.

\section{Conclusion} \label{sec:conclu}

Missing arbitrary modalities at inference time is expected in real-world applications, degrading the classification performance of multi-modal models.
In this work, we present two multi-modal co-learning methods to improve the classification robustness under inference scenarios with missing arbitrary modalities. 
Our methods learn modality-shared and -specific feature spaces through feature-based co-learning, while promoting predictive robustness via decision-level knowledge distillation.
Evaluated on two benchmarks, our methods consistently outperform (with significant differences) all competing approaches on one dataset, and achieve competitive (significantly similar) results in the other. 
The evidence highlights the potential and synergies of co-learning and knowledge distillation for enhancing the robustness of multi-modal classification models.


\begin{credits}
\subsubsection{\ackname} 
Funded by the Deutsche Forschungsgemeinschaft (DFG, German Research Foundation) – Project-ID 414984028 – SFB 1404 FONDA.



\end{credits}
%
%
%
\bibliographystyle{splncs04}
\bibliography{main}

\end{document}